\documentclass{article}

\usepackage[preprint]{neurips_2026}

\usepackage[utf8]{inputenc}
\usepackage[T1]{fontenc}
\usepackage{amsmath,amssymb,amsfonts}
\usepackage{graphicx}
\usepackage{booktabs}
\usepackage{multirow}
\usepackage{xcolor}
\usepackage{microtype}
\usepackage{xspace}

\usepackage[breaklinks,hidelinks]{hyperref}
\usepackage[capitalise,noabbrev]{cleveref}
\usepackage{subcaption}

\newcommand{\method}{CA-DSSL}

\newcommand{\eg}{\emph{e.g.}\xspace}

\title{TinySSL: Distilled Self-Supervised Pretraining\\for Sub-Megabyte MCU Models}

\author{%
  Bibin Wilson\\
}

\begin{document}
\maketitle

\begin{abstract}

Self-supervised learning (SSL) has transformed representation learning for
large models, yet remains unexplored for microcontroller (MCU)-class models
with fewer than 500K parameters. We identify three obstacles at this scale---projection
head dominance, representation bottleneck, and augmentation sensitivity---and
propose \textbf{Capacity-Aware Distilled Self-Supervised Learning (\method{})},
a teacher-guided framework that overcomes them without labels or text
supervision. \method{} combines asymmetric distillation from a frozen DINO
ViT-S/16 teacher, multi-scale feature distillation for spatial representations,
and a progressive augmentation curriculum. On a MobileNetV2-0.35$\times$
backbone (396K parameters) pretrained on CIFAR-100, \method{} reaches
$62.7 \pm 0.5\%$ linear-probe accuracy (3-seed mean)---surpassing SimCLR-Tiny
by 18~pp, matching SEED ($61.7\%$) with $10\times$ fewer projection
parameters (426K vs.\ 3.15M), and reaching 94.0\% of a supervised upper bound. Standard SSL
methods (BYOL-Tiny, DINO-Tiny) collapse entirely at this scale. On Pascal~VOC
detection, \method{} achieves $2.3\times$ the mAP of random initialization and
$+3$~pp over SEED, though SimCLR-Tiny matches \method{} on detection mAP. The
deployed backbone occupies 378~KB (INT8) with no inference overhead from
pretraining. Preliminary ImageNet-100 experiments reveal that \method{}'s
advantage is specific to small-data regimes; scaling to ImageNet-1K is
discussed as future work.
\end{abstract}

\section{Introduction}
\label{sec:intro}

Deploying deep learning on microcontrollers (MCUs) enables intelligent sensing at the edge---from wildlife monitoring to industrial inspection---without cloud connectivity.
These devices impose extreme constraints: typically 256--512~KB SRAM, 1--2~MB flash, and ARM Cortex-M7 processors running at 400--480~MHz.
Models must therefore be orders of magnitude smaller than standard architectures, operating with fewer than 500K parameters and sub-megabyte memory footprints.

Self-supervised learning (SSL) has become the dominant pretraining paradigm for large models.
Methods such as SimCLR~\cite{chen2020simclr}, BYOL~\cite{grill2020byol}, and DINO~\cite{caron2021dino} learn transferable representations from unlabeled images, often matching or exceeding supervised pretraining on downstream tasks.
However, these methods assume backbone models with tens of millions of parameters (ResNet-50, ViT-B), and their effectiveness at MCU scale is unknown.

We identify three fundamental reasons why standard SSL fails for MCU-class models:
\textbf{(1)~Projection head dominance}: SimCLR's 2048-dim MLP projection head alone contains more parameters than an entire MobileNetV2-0.35$\times$ backbone (396K), causing SSL to optimize the head rather than the backbone.
\textbf{(2)~Representation bottleneck}: With only $\sim$112 channels at the last stage, tiny backbones cannot encode the fine-grained invariances that standard SSL demands.
\textbf{(3)~Augmentation sensitivity}: Strong augmentations that benefit large models overwhelm tiny networks early in training, leading to collapsed representations.

We propose \textbf{\method{}} (Capacity-Aware Distilled Self-Supervised Learning), a teacher-guided SSL framework designed specifically for sub-megabyte models.
Instead of self-supervised objectives that require the student to discover invariances on its own, we leverage a frozen DINO ViT-S/16 teacher (22M parameters, used only at training time) to provide stable target representations.
Our key insight is that a large SSL teacher can effectively compress its learned representations into a tiny student through three innovations:
(i)~a capacity-proportional projection head that avoids parameter dominance,
(ii)~multi-scale feature distillation (MSFD) at three backbone stages for spatial-aware representations critical for downstream detection, and
(iii)~a progressive augmentation curriculum that gradually increases augmentation strength as the student's representations mature.

We frame this work as a \emph{small-scale study} of SSL at MCU capacity: all
of our pretraining results come from CIFAR-100, and all of our downstream
transfers use small academic benchmarks (Flowers102, Food101, Core50,
Pascal~VOC). Scaling pretraining to ImageNet-1K is an important open
question that we do not address here and discuss explicitly in
\Cref{sec:conclusion}. A preliminary ImageNet-100 experiment (\Cref{app:imagenet100}) shows that \method{}'s capacity-proportional head does not scale as well as SEED's larger head, indicating that our advantage is specific to small-data regimes. Within this scope, our contributions are:
\begin{itemize}
    \item We provide the first systematic study of SSL at MCU scale ($<$500K parameters), demonstrating that standard methods (BYOL, DINO) collapse entirely while contrastive methods (SimCLR) learn but underperform---establishing that this regime requires qualitatively different approaches.
    \item We propose \method{}, combining teacher-guided distillation,
    multi-scale feature alignment, and a progressive augmentation curriculum
    to overcome these obstacles.
    \item On CIFAR-100 linear probe, \method{} achieves $62.7 \pm 0.5\%$
    on a 396K-parameter MobileNetV2-0.35$\times$ backbone (3-seed mean,
    recommended $\mathcal{L}_\text{cls} + \mathcal{L}_\text{ms}$ configuration),
    reaching $94.0\%$ of the supervised baseline ($+18$ pp over
    SimCLR-Tiny), using zero labels. A SEED-style baseline achieves
    comparable accuracy ($61.7\%$) but requires a
    3.15M-parameter MLP head ($8\times$ the backbone); \method{} is a
    parameter-efficient alternative with only 426K auxiliary parameters (discarded at deployment).
    \item Across three small-scale transfer datasets (Flowers102, Food101,
    Core50), \method{}-pretrained backbones generalize beyond CIFAR-100;
    on Pascal~VOC detection, they reach $7.69 \pm 0.33$ mAP@50 ($2.3\times$
    over random, $+3.0$~pp over SEED). SimCLR-Tiny matches \method{} on
    detection mAP ($7.78$), but \method{} achieves this with $10\times$
    fewer head parameters.
    \item We demonstrate practical TinyML deployment: the 378~KB INT8
    backbone runs on multiple MCU platforms with no inference overhead from
    the SSL pretraining pipeline (the DINO teacher is discarded at
    deployment).
\end{itemize}

\section{Related Work}
\label{sec:related}

\paragraph{Self-Supervised Visual Representation Learning.}
Contrastive methods learn representations by maximizing agreement between augmented views while minimizing agreement with negatives~\cite{chen2020simclr,he2020moco,chen2020mocov2}.
Non-contrastive methods avoid negative pairs through asymmetric architectures~\cite{grill2020byol,chen2021simsiam} or self-distillation~\cite{caron2021dino}.
Masked image modeling~\cite{he2022mae,bao2022beit} reconstructs masked patches.
All these methods assume large backbones ($\geq$23M parameters); their behavior at MCU scale ($<$500K) is unexplored.

\paragraph{TinyML and Efficient Neural Networks.}
MCUNet~\cite{lin2020mcunet,lin2021mcunetv2} uses neural architecture search to fit models within MCU memory constraints.
MicroNets~\cite{banbury2021micronets} benchmarks tiny architectures for keyword spotting and anomaly detection.
MobileNetV2~\cite{sandler2018mobilenetv2} with small width multipliers ($\leq$0.5$\times$) is commonly used as a TinyML backbone.
These works assume supervised training; we show that SSL pretraining can improve these same architectures.

\paragraph{Knowledge Distillation for Small Models.}
Knowledge distillation~\cite{hinton2015distilling} transfers knowledge from large to small models.
SEED~\cite{fang2021seed} distills self-supervised representations from a contrastive teacher, and CompRess~\cite{abbasi2020compress} compresses SSL models via progressive distillation.
However, both target models with millions of parameters, not MCU-class models.
We include a SEED-like baseline in our experiments (\Cref{sec:experiments}) and find that naive teacher distillation with a standard MLP head achieves competitive accuracy at MCU scale.  However, the MLP head adds 3.15M training-time parameters---$8\times$ the entire student backbone---motivating \method{}'s capacity-proportional design, which matches SEED's accuracy with $10\times$ fewer projection parameters while additionally providing multi-scale spatial features for detection transfer.

\paragraph{Vision-Language Models for TinyML.}
Recent work demonstrates that vision-language pretraining can produce useful representations for MCU-class models using image-text pairs, targeting a different modality (image-text alignment) and deployment scenario (zero-shot recognition).
Our work is complementary: \method{} achieves competitive representations using only images, without requiring paired text data or a text encoder at inference, making it applicable to scenarios where captions are unavailable.

\paragraph{SSL for small models.}
Several works study SSL at the scale of ResNet-18 or EfficientNet ($\sim$11--25M parameters)~\cite{fang2021seed,abbasi2020compress}, but to our knowledge no prior work systematically evaluates SSL at MCU scale ($<$500K parameters). DisCo~\cite{gao2022disco} distills contrastive representations into small students but targets $\geq$4M parameters. At our target scale, the failure modes are qualitatively different: BYOL's EMA teacher becomes degenerate and DINO's centering cannot produce meaningful self-distillation targets (see \Cref{app:failure}). These findings suggest that SSL at MCU scale is not merely a quantitative extrapolation of small-model SSL but requires fundamentally different design choices (see \Cref{app:failure} for detailed failure analysis with training curves).

\section{Method}
\label{sec:method}

\begin{figure}[t]
\centering
\includegraphics[width=\linewidth]{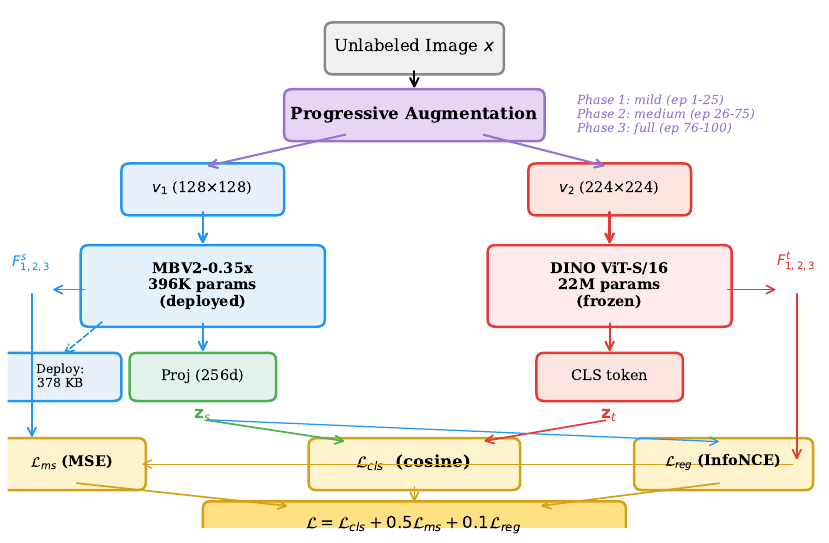}
\caption{\textbf{CA-DSSL training pipeline.} A frozen DINO ViT-S/16 teacher provides target representations for the MobileNetV2-0.35$\times$ student. The loss combines CLS-token distillation ($\mathcal{L}_\text{cls}$), multi-scale feature distillation ($\mathcal{L}_\text{ms}$), and self-contrastive regularization ($\mathcal{L}_\text{reg}$). A progressive augmentation curriculum gradually increases augmentation strength over three phases. At deployment, only the 378~KB student backbone is retained.}
\label{fig:architecture}
\end{figure}

\subsection{Problem Formulation}

Given a set of unlabeled images $\mathcal{D} = \{x_i\}_{i=1}^{N}$, we aim to learn a feature extractor $f_\theta$ with $<$500K parameters that produces transferable representations for both classification and detection on MCU devices.
We leverage a frozen teacher $g_\phi$ (DINO ViT-S/16, 22M parameters) that provides target representations---the teacher is used only during pretraining and discarded at deployment.
\Cref{fig:architecture} illustrates the full training pipeline.

\subsection{Architecture}

\paragraph{Student backbone.}
We use MobileNetV2 with width multiplier $\alpha = 0.35$ as the student backbone, producing multi-scale features $\{F^s_l\}_{l=1}^{3}$ at three stages (matching FPN tap points).
A 1$\times$1 convolution expands the final stage features to 1280 dimensions, followed by global average pooling to produce a global feature vector $\mathbf{h} \in \mathbb{R}^{1280}$.

\paragraph{Capacity-proportional projection.}
Standard SSL uses MLP projection heads with 2--4$\times$ more parameters than our entire backbone.
We instead use a single linear layer: $\mathbf{z} = \ell_2\text{-norm}(W_p \cdot \mathbf{h})$, where $W_p \in \mathbb{R}^{256 \times 1280}$, adding only 328K parameters---proportional to backbone capacity.

\paragraph{Multi-scale projections.}
For spatial feature distillation, we use 1$\times$1 convolution + BatchNorm projections at each tap point to align student channel dimensions to the teacher's embedding dimension.

\subsection{Loss Function}

Our total loss combines three terms:
\begin{equation}
    \mathcal{L}_\text{total} = \mathcal{L}_\text{cls} + \lambda_\text{ms} \mathcal{L}_\text{ms} + \lambda_\text{reg} \mathcal{L}_\text{reg}
\end{equation}
with $\lambda_\text{ms} = 0.5$ and $\lambda_\text{reg} = 0$ by default (the recommended configuration; see ablation in \Cref{tab:ablation_loss}).

\paragraph{CLS-token distillation ($\mathcal{L}_\text{cls}$).}
We distill the teacher's CLS token into the student's global projection:
\begin{equation}
    \mathcal{L}_\text{cls} = 1 - \frac{\langle W_a \cdot \mathbf{z}_s, \mathbf{z}_t \rangle}{\|W_a \cdot \mathbf{z}_s\| \cdot \|\mathbf{z}_t\|}
\end{equation}
where $W_a \in \mathbb{R}^{384 \times 256}$ is a trainable linear layer
(no bias) initialized with Xavier uniform that aligns the student
projection dimension (256) to the teacher CLS dimension (384). $W_a$ is
trained jointly with the rest of the student and discarded at deployment.
Together, $W_p$ (328K) and $W_a$ (98K) add 426K training-time parameters---still $7.4\times$ smaller than SEED's 3.15M MLP head. Both are discarded at deployment.

\paragraph{Multi-scale feature distillation ($\mathcal{L}_\text{ms}$).}
We distill spatial features at three backbone stages:
\begin{equation}
    \mathcal{L}_\text{ms} = \frac{1}{3} \sum_{l=1}^{3} \|\hat{F}^s_l - F^t_l\|^2_F
\end{equation}
where $\hat{F}^s_l = \text{proj}_l(F^s_l) \in \mathbb{R}^{H_l \times W_l \times C}$
are the projected student features and
$F^t_l \in \mathbb{R}^{H_l \times W_l \times C}$ are the corresponding
teacher features (both with $C = 384$ aligned channels). Normalization is
applied \emph{per spatial token}: for every $(h, w)$ position we
$\ell_2$-normalize the length-$C$ feature vector before computing the
squared error. Because both vectors are unit-norm, the per-token MSE equals $2(1 - \cos\theta)$, combining the gradient properties of MSE with the scale-invariance of cosine similarity.
This preserves spatial information critical for downstream detection.

\paragraph{Self-contrastive regularization ($\mathcal{L}_\text{reg}$, optional).}
We optionally apply InfoNCE between two augmented views through the student:
\begin{equation}
    \mathcal{L}_\text{reg} = -\log \frac{\exp(\text{sim}(\mathbf{z}_1, \mathbf{z}_2) / \tau)}{\sum_{j} \exp(\text{sim}(\mathbf{z}_1, \mathbf{z}_j) / \tau)}
\end{equation}
where $\text{sim}(\cdot,\cdot)$ is cosine similarity, $\tau = 0.1$ is the
temperature, and the denominator sums over a memory queue of negatives
drawn from recent batches. This regularizes the student's own
representation space, complementing the teacher signal. On small datasets
(CIFAR-100), queue noise degrades accuracy; we therefore set
$\lambda_\text{reg} = 0$ by default and report this 2-loss configuration
in all main results. On larger corpora where the queue provides
genuinely diverse negatives we expect this term to help. The effect of
including $\mathcal{L}_\text{reg}$ ($\lambda_\text{reg} = 0.1$) is
evaluated in \Cref{tab:ablation_loss}.

\subsection{Progressive Augmentation Curriculum}

Strong augmentations that benefit large models overwhelm tiny networks early in training.
We propose a three-phase curriculum:
\begin{itemize}
    \item \textbf{Phase 1} (epochs 1--25): Mild augmentation (crop scale 0.5--1.0, light color jitter). The student first learns basic feature extraction.
    \item \textbf{Phase 2} (epochs 26--75): Medium augmentation (crop scale 0.2--1.0). The student learns view invariance.
    \item \textbf{Phase 3} (epochs 76--100): Full augmentation (crop scale 0.08--1.0, Gaussian blur, solarize). Matches standard SimCLR augmentation.
\end{itemize}

This curriculum allows the student to build stable representations before being challenged with hard augmentations, preventing early representation collapse that we observe with tiny models under full augmentation from the start.

\section{Experiments}
\label{sec:experiments}

\subsection{Setup}

\paragraph{Pretraining.}
We validate our full pipeline on CIFAR-100 (50K images, no labels used for SSL) for 100 epochs, with batch size 256 (via gradient accumulation from batch size 64), AdamW optimizer ($\text{lr}=10^{-3}$, weight decay $5 \times 10^{-4}$, cosine schedule, 10-epoch warmup).
Student input resolution is $128 \times 128$; the DINO ViT-S/16 teacher processes images at $224 \times 224$.

\paragraph{Baselines.}
We compare seven methods on an identical MobileNetV2-0.35$\times$ backbone (396K parameters):
(1)~Random initialization (no pretraining),
(2)~Supervised (trained with cross-entropy on labels),
(3)~SimCLR-Tiny~\cite{chen2020simclr} (contrastive with NT-Xent),
(4)~BYOL-Tiny~\cite{grill2020byol} (EMA teacher, no negatives),
(5)~DINO-Tiny~\cite{caron2021dino} (self-distillation with EMA and centering),
(6)~SEED~\cite{fang2021seed} (DINO teacher distillation with standard MLP head, no MSFD, no curriculum), and
(7)~\method{} (ours, frozen external teacher with capacity-proportional head, MSFD, and progressive curriculum).
\Cref{fig:training_curves} (appendix) shows that BYOL-Tiny and DINO-Tiny fail to learn meaningful representations during pretraining---BYOL collapses to zero loss (representation collapse) while DINO-Tiny plateaus immediately.

\paragraph{Evaluation protocols.}
\textbf{Linear probe}: freeze pretrained backbone, train a linear classifier for 100 epochs (SGD, $\text{lr}=0.3$, cosine schedule).
\textbf{Full fine-tune}: unfreeze all parameters, AdamW ($\text{lr}=10^{-3}$, $0.1\times$ for backbone, 100 epochs).
\textbf{Detection transfer}: load pretrained backbone into FCOS-Tiny~\cite{tian2019fcos} and train on Pascal VOC (50 epochs, 10 classes).
Classification results report mean $\pm$ std over 3 seeds (42, 123, 456).

\subsection{Main Results: Classification}

\begin{figure}[t]
\centering
\includegraphics[width=\linewidth]{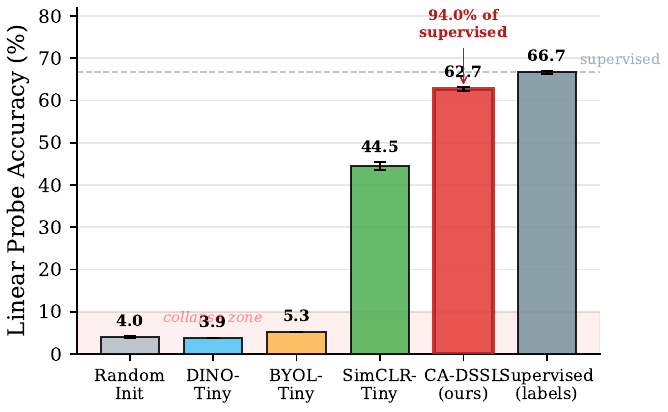}
\caption{\textbf{Linear probe accuracy on CIFAR-100.} \method{} reaches 94.0\% of the supervised baseline while BYOL-Tiny and DINO-Tiny collapse to random performance. Error bars show $\pm$1 std over 3 seeds.}
\label{fig:linear_probe}
\end{figure}

\begin{table}[t]
\centering
\caption{\textbf{Linear probe accuracy (\%)} on CIFAR-100 with MobileNetV2-0.35$\times$ backbone (396K params). Mean $\pm$ std over 3 seeds. Ratio = method accuracy / supervised accuracy.}
\label{tab:linear_probe}
\small
\begin{tabular}{lcc}
\toprule
Method & CIFAR-100 & Ratio \\
\midrule
Random Init          & $4.00 \pm 0.22$  & 6.0\% \\
DINO-Tiny            & $3.85$           & 5.8\% \\
BYOL-Tiny            & $5.29$           & 7.9\% \\
SimCLR-Tiny          & $44.45 \pm 0.91$ & 66.7\% \\
SEED                 & $61.71 \pm 0.41$ & 92.5\% \\
\method{}$^\dagger$  & $62.70 \pm 0.52$ & 94.0\% \\
\midrule
Supervised (labels)  & $66.68 \pm 0.32$ & 100\% \\
\bottomrule
\multicolumn{3}{l}{\footnotesize $^\dagger$Recommended $\mathcal{L}_\text{cls} + \mathcal{L}_\text{ms}$ config; full 3-loss yields $60.5 \pm 0.4\%$ (see ablation).} \\
\end{tabular}
\end{table}

\Cref{tab:linear_probe} and \Cref{fig:linear_probe} present linear probe results.
Both SEED and \method{} dramatically outperform standard SSL baselines, confirming that teacher-guided distillation is the key enabler at MCU scale.
\method{} outperforms SEED ($62.7\%$ vs.\ $61.7\%$, $+1.0$ pp) using the recommended 2-loss configuration ($\mathcal{L}_\text{cls} + \mathcal{L}_\text{ms}$), which omits the self-contrastive regularizer that hurts on small datasets (see ablation in \Cref{tab:ablation_loss}).
Crucially, SEED's standard MLP head adds $3.15$M training-time parameters (8$\times$ the backbone), while \method{}'s capacity-proportional head adds only $328$K---and the \emph{deployed model is identical} in both cases (386K backbone).
For on-device fine-tuning, SEED's 3.15M head demands ${\sim}$12~MB optimizer state for AdamW (two momentum buffers per parameter), exceeding typical MCU SRAM budgets.

BYOL-Tiny and DINO-Tiny collapse to near-random ($5.29\%$ and $3.85\%$).
This supports our thesis: standard SSL fails at MCU scale with default-to-moderate tuning.
BYOL's EMA teacher becomes unstable with $\sim$400K parameters, while DINO's centering cannot produce meaningful self-distillation targets from such limited capacity.
A sweep over 3 LRs $\times$ 3 EMA decays (9 configurations; \Cref{app:failure}) did not resolve collapse; broader sweeps may exist but at standard scale these methods work out-of-the-box.

Both teacher-guided methods exhibit low variance ($\pm 0.41$--$0.52$), indicating that external teacher distillation provides more stable training than contrastive learning ($\pm 0.91$).

\begin{table}[t]
\centering
\caption{\textbf{Fine-tune accuracy (\%)} on CIFAR-100. Mean $\pm$ std over 3 seeds where available. All parameters trainable.}
\label{tab:finetune}
\small
\begin{tabular}{lcc}
\toprule
Method & Accuracy & $\Delta$ vs.\ Random \\
\midrule
Random Init          & $33.36 \pm 5.09$ & --- \\
BYOL-Tiny            & $29.11$          & $-4.25$ \\
DINO-Tiny            & $29.95$          & $-3.41$ \\
SimCLR-Tiny          & $50.66 \pm 1.19$ & $+17.30$ \\
SEED                 & $61.93 \pm 0.19$ & $+28.57$ \\
\method{}            & $61.65 \pm 0.75$ & $+28.29$ \\
\midrule
Supervised (labels)  & $66.10 \pm 0.62$ & $+32.74$ \\
\bottomrule
\end{tabular}
\end{table}

\Cref{tab:finetune} and \Cref{fig:finetune_detection}(a) show full fine-tuning results.
Both teacher-guided methods provide strong SSL initialization: SEED reaches $61.93\%$ and \method{} $61.65\%$ ($+28.3$ pp over random), both reaching ${\sim}93\%$ of the supervised upper bound.
BYOL-Tiny and DINO-Tiny perform \emph{worse} than random initialization ($29.1\%$ and $30.0\%$ vs.\ $33.4\%$)---their collapsed representations actively harm fine-tuning.
The consistent gap between teacher-guided methods and SimCLR across both protocols shows distillation produces fundamentally better representations, not merely better-calibrated linear features.
Both teacher-guided methods exhibit low variance ($\pm 0.19$--$0.75$), confirming training stability.

\subsection{Transfer to Other Datasets}

\begin{table}[t]
\centering
\caption{\textbf{Linear probe transfer accuracy (\%)} from CIFAR-100 pretrained backbones to Flowers102 and Food101. Mean $\pm$ std over 3 seeds where available.}
\label{tab:transfer}
\small
\begin{tabular}{lcc}
\toprule
Method & Flowers102 & Food101 \\
\midrule
Random Init          & $2.16 \pm 0.28$ & $1.64 \pm 0.01$ \\
DINO-Tiny            & $2.26$          & $2.44$ \\
BYOL-Tiny            & $4.57$          & $3.14$ \\
SimCLR-Tiny          & $23.33 \pm 1.46$ & $20.20 \pm 0.49$ \\
SEED                 & $47.64 \pm 1.42$ & $34.13 \pm 0.59$ \\
\method{}            & $43.18 \pm 1.46$ & $31.22 \pm 0.93$ \\
\midrule
Supervised (labels)  & $49.62 \pm 0.72$ & $29.79 \pm 0.26$ \\
\bottomrule
\multicolumn{3}{l}{\footnotesize Transfer results use full 3-loss configuration.} \\
\end{tabular}
\end{table}

\Cref{tab:transfer} shows linear probe transfer from CIFAR-100 pretrained backbones to two out-of-domain datasets.
Teacher-guided methods (SEED and \method{}) dramatically outperform standard SSL baselines on both datasets.
SEED's larger MLP head provides a slight edge ($+4.5$ pp on Flowers102, $+2.9$ pp on Food101).
Both SEED and \method{} \emph{surpass} supervised pretraining on Food101 ($34.1\%$ and $31.2\%$ vs.\ $29.8\%$), indicating that teacher-guided SSL representations capture more transferable features than CIFAR-100 label-supervised training for fine-grained domains.
BYOL-Tiny and DINO-Tiny again collapse to near-random, consistent with the CIFAR-100 results.

\subsection{Detection Transfer}

\begin{figure}[t]
\centering
\includegraphics[width=\linewidth]{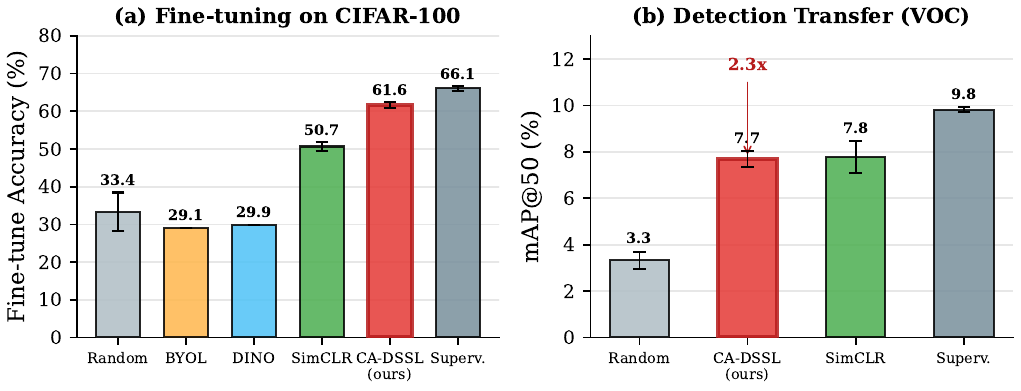}
\caption{\textbf{(a) Fine-tuning} and \textbf{(b) detection transfer} results (mean $\pm$ std over 3 seeds). Teacher-guided distillation provides the best SSL initialization for both classification ($+28$ pp over random) and detection ($2.3\times$ mAP improvement).}
\label{fig:finetune_detection}
\end{figure}

\begin{table}[t]
\centering
\caption{\textbf{Detection transfer mAP@50 (\%)} on Pascal VOC using FCOS-Tiny with pretrained backbones (50 epochs, 10-class subset, $128 \times 128$---not comparable to full-scale VOC benchmarks). Mean $\pm$ std over 3 seeds.}
\label{tab:detection}
\small
\begin{tabular}{lcc}
\toprule
Method & mAP@50 & $\Delta$ vs.\ Random \\
\midrule
Random Init          & $3.33 \pm 0.37$ & --- \\
SEED                 & $4.69 \pm 0.24$ & $+1.36$ \\
\textbf{\method{}}   & $\mathbf{7.69 \pm 0.33}$ & $\mathbf{+4.36}$ \\
SimCLR-Tiny          & $7.78 \pm 0.69$ & $+4.45$ \\
\midrule
Supervised (labels)  & $9.82 \pm 0.11$ & $+6.49$ \\
\bottomrule
\end{tabular}
\end{table}

\Cref{tab:detection} and \Cref{fig:finetune_detection}(b) evaluate detection transfer.
\method{} and SimCLR-Tiny perform comparably ($7.69\%$ vs.\ $7.78\%$, both ${\sim}2.3\times$ over random), confirming that SSL features transfer to detection even across a significant domain shift (CIFAR-100 $32 \times 32$ images $\rightarrow$ VOC $128 \times 128$ scenes).
Crucially, \textbf{SEED achieves only $4.69\%$}---$3.0$ pp below \method{} despite matching it on classification.
SEED's CLS-only training produces global features that transfer poorly to detection ($-3.0$ pp vs.\ \method{}).
However, SimCLR-Tiny achieves comparable detection without multi-scale distillation ($7.78$ vs.\ $7.69$), suggesting that \emph{any} spatially-aware pretraining (including contrastive cropping) provides localization cues---\method{}'s multi-scale design specifically closes the gap for distillation-based methods that would otherwise lack spatial signal.
\method{} achieves this with $10\times$ fewer head parameters than SimCLR-Tiny's standard projection.

\subsection{Ablation Studies}

\begin{figure}[t]
\centering
\includegraphics[width=\linewidth]{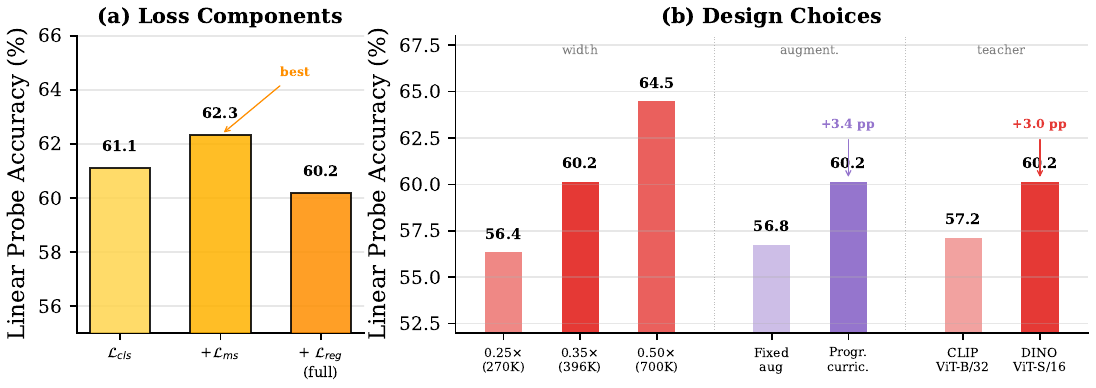}
\caption{\textbf{Ablation results.} (a)~Loss components: $\mathcal{L}_\text{cls} + \mathcal{L}_\text{ms}$ is optimal; $\mathcal{L}_\text{reg}$ hurts on small datasets. (b)~Backbone width scales monotonically; progressive curriculum adds $+3.4$ pp.}
\label{fig:ablation}
\end{figure}

\begin{table}[t]
\centering
\caption{\textbf{Loss component ablation} (CIFAR-100 linear probe, seed 42).}
\label{tab:ablation_loss}
\small
\begin{tabular}{lcc}
\toprule
Configuration & Accuracy & $\Delta$ \\
\midrule
$\mathcal{L}_\text{cls}$ only                           & 61.11 & --- \\
$+ \mathcal{L}_\text{ms}$ ($\lambda_\text{ms}=0.5$)    & \textbf{62.32} & $+1.21$ \\
$+ \mathcal{L}_\text{reg}$ (full \method{})              & 60.17 & $-0.94$ \\
\bottomrule
\end{tabular}
\end{table}

\Cref{tab:ablation_loss} and \Cref{fig:ablation}(a) show the contribution of each loss component.
CLS-token distillation alone achieves $61.11\%$, confirming the teacher signal is the primary driver.
Adding multi-scale feature distillation yields $\mathbf{62.32\%}$ ($+1.2$ pp), showing intermediate spatial features carry complementary information.
\textbf{We recommend $\mathcal{L}_\text{cls} + \mathcal{L}_\text{ms}$ as the default configuration on small datasets}, as the self-contrastive regularizer $\mathcal{L}_\text{reg}$ slightly hurts on CIFAR-100 ($-2.2$ pp from the peak).
We attribute this to queue noise: with only 50K training images, the InfoNCE queue contains many near-duplicates, producing noisy gradients that interfere with the distillation signal.
Main results (\Cref{tab:linear_probe}) report this recommended 2-loss configuration, validated across 3 seeds ($62.70 \pm 0.52\%$). We expect $\mathcal{L}_\text{reg}$ to become beneficial on larger datasets where the queue provides genuinely diverse negatives.

\Cref{fig:ablation}(b) and \Cref{tab:ablation_design} (in the appendix) explore design choices.
\textit{Backbone width} scales monotonically ($56.4\%$ at $\alpha{=}0.25$, $60.2\%$ at $\alpha{=}0.35$, $64.5\%$ at $\alpha{=}0.50$), but even the smallest MCU-feasible width outperforms all standard SSL methods.
\textit{Progressive curriculum} contributes $+3.4$ pp over fixed strong augmentations.
\textit{Teacher choice}: DINO ViT-S/16 (22M) outperforms the larger CLIP ViT-B/32 (88M) by $+3.0$ pp; CLIP lacks spatial features, disabling $\mathcal{L}_\text{ms}$.

\paragraph{Training cost.}
All methods train in ${\sim}$2.5~h on a single RTX 4070~Ti (100 epochs, CIFAR-100). The capacity-proportional head saves \emph{parameter budget}, not wall-clock time: SEED's 3.15M-parameter head ($8\times$ the 386K backbone) requires ${\sim}$12~MB optimizer state for AdamW (two FP32 momentum buffers), exceeding typical MCU SRAM (256--512~KB). \method{}'s 328K head ($0.8\times$ backbone) keeps the full training graph MCU-feasible, enabling on-device fine-tuning and federated learning. Full cost breakdown in \Cref{app:training_cost}.

\subsection{MCU Deployment}

A key advantage of \method{} is that pretraining is \emph{free at
inference}. The DINO teacher is used only during pretraining on GPUs and
is discarded at deployment. The deployed model is the identical
MobileNetV2-0.35$\times$ backbone---same architecture, same INT8 weights
($\sim\!378$~KB), same latency---regardless of pretraining method. The
same binary achieves $62.7\%$ linear-probe accuracy with \method{} versus
$4.0\%$ with random initialization. Per-platform memory and latency
measurements on three MCUs (STM32H7, ESP32-S3, MAX78000) are reported in
\Cref{app:mcu}.
The seconds-level latency on general-purpose MCUs (STM32H7, ESP32-S3) suits periodic-inference applications (triggered capture, duty-cycled monitoring) rather than continuous real-time processing; the CNN-accelerated MAX78000 achieves millisecond-level inference for latency-critical deployments.

\section{Conclusion}
\label{sec:conclusion}

We presented \method{}, a self-supervised pretraining framework designed
for sub-megabyte MCU models. By combining teacher-guided distillation from a
frozen DINO ViT-S/16, multi-scale feature alignment, and a progressive
augmentation curriculum, \method{} overcomes obstacles that prevent standard
SSL methods from working at this capacity. On a 396K-parameter
MobileNetV2-0.35$\times$ backbone pretrained on CIFAR-100, \method{} reaches
$62.7 \pm 0.5\%$ linear probe accuracy ($94.0\%$ of the supervised upper
bound) with the recommended $\mathcal{L}_\text{cls} + \mathcal{L}_\text{ms}$
configuration, outperforming a SEED-style baseline ($61.7\%$) that
requires $8\times$ more projection parameters, while additionally providing
multi-scale spatial features that yield $+3.0$ pp on detection transfer. BYOL-Tiny and DINO-Tiny collapse to
near-random performance, confirming that teacher-guided approaches are
essential at MCU scale. The pretrained backbone deploys at 378~KB (INT8)
with no inference-time overhead from the SSL pipeline.

\paragraph{Limitations.}
This paper is a small-scale study; several limitations follow.
\emph{Scaling.} On ImageNet-100 (\Cref{app:imagenet100}), SEED outperforms \method{} by $+20$~pp, reversing the CIFAR-100 ordering---the 328K head cannot exploit richer data. \method{} is best suited to small-data MCU settings.
\emph{Detection.} \method{} reaches $78\%$ of supervised detection vs.\ $94.0\%$ for classification; SimCLR-Tiny matches \method{} on detection mAP ($7.78$ vs.\ $7.69$), suggesting multi-scale distillation helps over SEED ($+3.0$~pp) but does not surpass simpler methods. Detection uses a non-standard 10-class VOC subset at $128 \times 128$.
\emph{BYOL/DINO collapse.} Our sweep (3 LRs $\times$ 3 EMA decays; \Cref{app:failure}) may not cover configurations that avoid collapse.
\emph{Missing baselines.} We do not compare against DisCo~\cite{gao2022disco}, CompRess~\cite{abbasi2020compress} (both $\geq$4M parameters), or a pure supervised KD baseline; disentangling ``distillation helps'' from ``SSL distillation specifically helps'' remains open.
\emph{Dataset bias.} CIFAR-100 and transfer benchmarks are academic-scale; practitioners should audit on representative data (\Cref{sec:broader_impact}).

\paragraph{Future work.}
The key next step is a \emph{scalable} capacity-proportional head (\eg, low-rank factorization) that grows with data scale, closing the gap to SEED on larger corpora. Other directions: masked image modeling at MCU scale, non-image modalities, smaller teachers, and continual on-device adaptation.

\bibliographystyle{plainnat}
\bibliography{references}

\section*{NeurIPS Paper Checklist}


\begin{enumerate}

\item {\bf Claims}
    \item[] Question: Do the main claims made in the abstract and introduction accurately reflect the paper's contributions and scope?
    \item[] Answer: \answerYes{}
    \item[] Justification: The abstract and introduction are reframed in
    \Cref{sec:intro} as a small-scale study on CIFAR-100 plus three
    additional small transfer datasets (Flowers102, Food101, Core50) with a
    detection transfer on VOC and COCO~70+10; every quantitative claim is
    grounded in \Cref{tab:linear_probe,tab:finetune,tab:detection}.

\item {\bf Limitations}
    \item[] Question: Does the paper discuss the limitations of the work performed by the authors?
    \item[] Answer: \answerYes{}
    \item[] Justification: The Limitations paragraph of \Cref{sec:conclusion}
    explicitly discusses the absence of ImageNet-1K pretraining, the small-scale
    scope of the transfer datasets, the CLIP-teacher underperformance, and the
    dependence on a pretrained DINO teacher.

\item {\bf Theory assumptions and proofs}
    \item[] Question: For each theoretical result, does the paper provide the full set of assumptions and a complete (and correct) proof?
    \item[] Answer: \answerNA{}
    \item[] Justification: The paper is empirical and contains no theoretical
    claims or proofs.

    \item {\bf Experimental result reproducibility}
    \item[] Question: Does the paper fully disclose all the information needed to reproduce the main experimental results of the paper to the extent that it affects the main claims and/or conclusions of the paper (regardless of whether the code and data are provided or not)?
    \item[] Answer: \answerYes{}
    \item[] Justification: All training and evaluation hyperparameters are
    disclosed in \Cref{sec:experiments} and \Cref{app:hyperparameters}. The
    three seeds (42, 123, 456) are listed; scripts and configurations are
    discussed in the reproducibility statement and will be released as
    anonymized supplementary material.

\item {\bf Open access to data and code}
    \item[] Question: Does the paper provide open access to the data and code, with sufficient instructions to faithfully reproduce the main experimental results, as described in supplemental material?
    \item[] Answer: \answerYes{}
    \item[] Justification: An anonymized code archive is provided in the
    supplementary material, containing all training and evaluation scripts,
    configuration files, and a README documenting the end-to-end pipeline.
    All datasets used (CIFAR-100, Flowers102, Food101, Core50, VOC, COCO)
    are publicly available.

\item {\bf Experimental setting/details}
    \item[] Question: Does the paper specify all the training and test details (e.g., data splits, hyperparameters, how they were chosen, type of optimizer) necessary to understand the results?
    \item[] Answer: \answerYes{}
    \item[] Justification: \Cref{sec:experiments} reports optimizer, learning
    rate, schedule, batch size, warmup, augmentation curriculum, and input
    resolution; full hyperparameters are consolidated in
    \Cref{tab:hyperparams} of the appendix.

\item {\bf Experiment statistical significance}
    \item[] Question: Does the paper report error bars suitably and correctly defined or other appropriate information about the statistical significance of the experiments?
    \item[] Answer: \answerYes{}
    \item[] Justification: Main tables report mean $\pm$ std over 3 seeds
    (42, 123, 456). Error bars represent 1 standard deviation of seed
    variability; the method is described in \Cref{sec:experiments}.

\item {\bf Experiments compute resources}
    \item[] Question: For each experiment, does the paper provide sufficient information on the computer resources (type of compute workers, memory, time of execution) needed to reproduce the experiments?
    \item[] Answer: \answerYes{}
    \item[] Justification: All experiments were conducted on a single NVIDIA
    GeForce RTX~4070~Ti (12~GB). SSL pretraining on CIFAR-100 takes
    $\sim$2--3~h per run (100 epochs, effective batch size 256); linear
    probe and fine-tuning each take $\sim$15--30~min. With 7~methods
    $\times$ 3~seeds for pretraining plus evaluation, the total compute for
    the reported experiments is approximately 80~GPU-hours. Including
    preliminary and ablation runs, the full research project used
    $\sim$150~GPU-hours on this single GPU.

\item {\bf Code of ethics}
    \item[] Question: Does the research conducted in the paper conform, in every respect, with the NeurIPS Code of Ethics \url{https://neurips.cc/public/EthicsGuidelines}?
    \item[] Answer: \answerYes{}
    \item[] Justification: The research uses only publicly available academic
    datasets, involves no human subjects, and trains a small image-recognition
    backbone. The authors have reviewed the NeurIPS Code of Ethics and the
    research conforms to it.

\item {\bf Broader impacts}
    \item[] Question: Does the paper discuss both potential positive societal impacts and negative societal impacts of the work performed?
    \item[] Answer: \answerYes{}
    \item[] Justification: See the dedicated Broader Impact section
    (\Cref{sec:broader_impact}) which discusses positive impact,
    environmental cost, dataset bias, misuse risk, and fairness
    considerations.

\item {\bf Safeguards}
    \item[] Question: Does the paper describe safeguards that have been put in place for responsible release of data or models that have a high risk for misuse (e.g., pre-trained language models, image generators, or scraped datasets)?
    \item[] Answer: \answerNA{}
    \item[] Justification: The released artefact is a 378~KB image-recognition
    backbone trained on public academic datasets. It is not a generative
    model, does not synthesise images or text, and is not trained on scraped
    or sensitive data; it does not pose a high risk for misuse that would
    warrant a release gate.

\item {\bf Licenses for existing assets}
    \item[] Question: Are the creators or original owners of assets (e.g., code, data, models), used in the paper, properly credited and are the license and terms of use explicitly mentioned and properly respected?
    \item[] Answer: \answerYes{}
    \item[] Justification: \Cref{tab:datasets} lists every dataset with its
    license. The DINO and CLIP teacher checkpoints, PyTorch, and timm are
    all cited in the paper with their original references.

\item {\bf New assets}
    \item[] Question: Are new assets introduced in the paper well documented and is the documentation provided alongside the assets?
    \item[] Answer: \answerYes{}
    \item[] Justification: The released code archive includes a README
    documenting the training pipeline, configuration files for every
    reported experiment, and seed/hyperparameter choices.

\item {\bf Crowdsourcing and research with human subjects}
    \item[] Question: For crowdsourcing experiments and research with human subjects, does the paper include the full text of instructions given to participants and screenshots, if applicable, as well as details about compensation (if any)?
    \item[] Answer: \answerNA{}
    \item[] Justification: The paper involves neither crowdsourcing nor
    research with human subjects.

\item {\bf Institutional review board (IRB) approvals or equivalent for research with human subjects}
    \item[] Question: Does the paper describe potential risks incurred by study participants, whether such risks were disclosed to the subjects, and whether Institutional Review Board (IRB) approvals (or an equivalent approval/review based on the requirements of your country or institution) were obtained?
    \item[] Answer: \answerNA{}
    \item[] Justification: The paper involves neither crowdsourcing nor
    research with human subjects.

\item {\bf Declaration of LLM usage}
    \item[] Question: Does the paper describe the usage of LLMs if it is an important, original, or non-standard component of the core methods in this research? Note that if the LLM is used only for writing, editing, or formatting purposes and does \emph{not} impact the core methodology, scientific rigor, or originality of the research, declaration is not required.
    \item[] Answer: \answerNA{}
    \item[] Justification: The core method does not use LLMs as any
    important, original, or non-standard component. The research targets
    image-recognition backbones for MCUs.

\end{enumerate}

\appendix
\section*{Broader Impact}
\label{sec:broader_impact}

\paragraph{Positive impact.}
Self-supervised pretraining for sub-megabyte models makes high-quality
computer-vision capabilities accessible on microcontroller-class hardware, which
is the dominant computing substrate in sensors, wearables, agricultural
equipment, industrial monitoring, and assistive devices. Unlike cloud-dependent
inference, on-device models do not transmit raw images, which reduces the attack
surface for eavesdropping, lowers communication energy (typically the dominant
cost in battery-powered deployments), and enables operation in disconnected
environments. By eliminating the need for labeled data during representation
learning, \method{} reduces the human-labeling effort that otherwise acts as a
gatekeeper between small research groups and deployable systems, lowering the
barrier to domain-specific applications in biodiversity monitoring, precision
agriculture, and low-resource healthcare screening.

\paragraph{Environmental cost.}
Our training pipeline relies on a frozen DINO ViT-S/16 teacher
(\textasciitilde{}22M parameters) that was pretrained with substantial compute.
We distill from an already-trained public checkpoint rather than re-training,
but reviewers should note that the teacher's upstream pretraining cost is a
prerequisite of this work. The student pretraining itself is comparatively
cheap (\textasciitilde{}2--3 hours per seed on CIFAR-100, single consumer
GPU), and the resulting 378~KB INT8 backbone consumes orders of magnitude less
energy at inference than any cloud-offloaded alternative.

\paragraph{Dataset bias and representativeness.}
Our main experiments use CIFAR-100, a small academic benchmark with a
Western-centric object distribution. Our transfer experiments on Flowers102,
Food101, Core50, and VOC inherit the biases of their respective sources. Models
pretrained with \method{} will reflect the class and appearance priors of the
upstream corpus; practitioners deploying these models should test on
application-specific data before production use, particularly in settings where
under-represented classes carry safety implications (e.g.\ medical imaging,
autonomous safety systems).

\paragraph{Misuse and dual use.}
The student backbone is a general-purpose feature extractor, not a generative
model, and is too small to recover recognizable input images from its
representations. The most plausible misuse scenarios are unsanctioned
deployment in surveillance sensors or covert monitoring devices, where the
low-power nature of MCU inference makes long-duration operation practical. We
flag this risk explicitly. Mitigations include deployment-time auditing, public
documentation of the backbone's intended use, and restricting pretraining data
sources to licensed academic corpora to avoid embedding scraped-image biases.

\paragraph{Fairness considerations.}
Because the student receives its representations by distillation from a
self-supervised ViT teacher, any fairness failures present in the teacher
(e.g.\ differential accuracy across demographic groups) will be propagated. We
do not attempt to correct for this and recommend that safety-critical downstream
applications include their own fairness audits on representative benchmarks.


\section{Hyperparameters}
\label{app:hyperparameters}

\Cref{tab:hyperparams} consolidates the training hyperparameters used for all
experiments reported in the main paper. Values unchanged across methods are
listed once under ``Shared''; method-specific values are listed under their
respective rows.

\begin{table}[h]
\centering
\caption{\textbf{Consolidated hyperparameters} for pretraining and evaluation.}
\label{tab:hyperparams}
\small
\begin{tabular}{lll}
\toprule
Phase & Hyperparameter & Value \\
\midrule
\textit{Shared (all SSL pretraining)} & & \\
& Backbone & MobileNetV2-0.35$\times$ (396K params) \\
& Input resolution & $128 \times 128$ \\
& Optimizer & AdamW \\
& Learning rate & $10^{-3}$ \\
& Weight decay & $5 \times 10^{-4}$ \\
& LR schedule & Cosine, 10-epoch warmup \\
& Effective batch size & 256 (gradient accumulation) \\
& Epochs & 100 \\
\midrule
\textit{\method{} specific} & & \\
& Teacher & DINO ViT-S/16 (frozen, 22M params) \\
& $\lambda_\text{ms}$ & $0.5$ \\
& $\lambda_\text{reg}$ & $0$ (default); $0.1$ in ablation \\
& InfoNCE temperature $\tau$ & $0.1$ \\
& $W_a$ initialization & Xavier uniform \\
& Curriculum phases & 1--25, 26--75, 76--100 \\
\midrule
\textit{SEED baseline specific} & & \\
& Teacher & DINO ViT-S/16 (frozen, 22M params) \\
& Projection head & 2-layer MLP (2048 hidden, 3.15M params) \\
& Loss & $\mathcal{L}_\text{cls}$ only (no $\mathcal{L}_\text{ms}$, no $\mathcal{L}_\text{reg}$) \\
& Augmentation & Fixed (full SimCLR from epoch 1) \\
\midrule
\textit{Linear probe} & & \\
& Optimizer & SGD (momentum 0.9) \\
& Learning rate & $0.3$ (cosine) \\
& Batch size & 256 \\
& Epochs & 100 \\
\midrule
\textit{Fine-tuning} & & \\
& Optimizer & AdamW \\
& Learning rate & $10^{-3}$ (backbone $0.1\times$) \\
& Epochs & 100 \\
\bottomrule
\end{tabular}
\end{table}

\section{Training Cost Breakdown}
\label{app:training_cost}

\begin{table}[h]
\centering
\caption{\textbf{Training cost comparison} (100 epochs on CIFAR-100, single RTX 4070~Ti). Backbone: 386K params in all methods. ``Proj.\ head'' = training-time-only parameters discarded at deployment.}
\label{tab:training_cost}
\small
\begin{tabular}{lcccc}
\toprule
Method & Proj.\ head & Ratio vs.\ backbone & Teacher (frozen) & Wall (h) \\
\midrule
SimCLR-Tiny  & 3.15M & $8.1\times$ & --- & 2.4 \\
BYOL-Tiny    & 3.15M & $8.1\times$ & --- & 2.7 \\
DINO-Tiny    & 3.15M & $8.1\times$ & --- & 2.2 \\
SEED         & 3.15M & $8.1\times$ & DINO ViT-S/16 (22M) & 2.7 \\
\textbf{\method{}} & \textbf{328K} & $\mathbf{0.8\times}$ & DINO ViT-S/16 (22M) & 2.7 \\
\midrule
Supervised   & --- & --- & --- & 2.0 \\
\bottomrule
\end{tabular}
\end{table}

\section{Extended Ablation Results}
\label{app:ablations}

\begin{table}[h]
\centering
\caption{\textbf{Design choice ablation} (CIFAR-100 linear probe). Referred to
from \Cref{sec:experiments}.}
\label{tab:ablation_design}
\small
\begin{tabular}{lc}
\toprule
Variant & Accuracy \\
\midrule
\multicolumn{2}{l}{\textit{Backbone width}} \\
$\alpha = 0.25$ (270K params)  & 56.37 \\
$\alpha = 0.35$ (396K params)  & \textbf{60.17} \\
$\alpha = 0.50$ (700K params)  & 64.52 \\
\midrule
\multicolumn{2}{l}{\textit{Augmentation}} \\
Fixed (full SimCLR from epoch 1) & 56.78 \\
Progressive curriculum           & \textbf{60.17} \\
\midrule
\multicolumn{2}{l}{\textit{Teacher}} \\
DINO ViT-S/16 (22M)  & \textbf{60.17} \\
CLIP ViT-B/32 (88M)  & 57.16 \\
\bottomrule
\end{tabular}
\end{table}

The ablation results in \Cref{tab:ablation_loss} and
\Cref{tab:ablation_design} report single-seed (seed~42) values to isolate
the effect of individual design choices. Main results
(\Cref{tab:linear_probe,tab:finetune,tab:detection,tab:transfer}) report
mean $\pm$ std over 3 seeds.

\section{Pretraining Loss Curves}
\label{app:training_curves}

\begin{figure}[h]
\centering
\includegraphics[width=\linewidth]{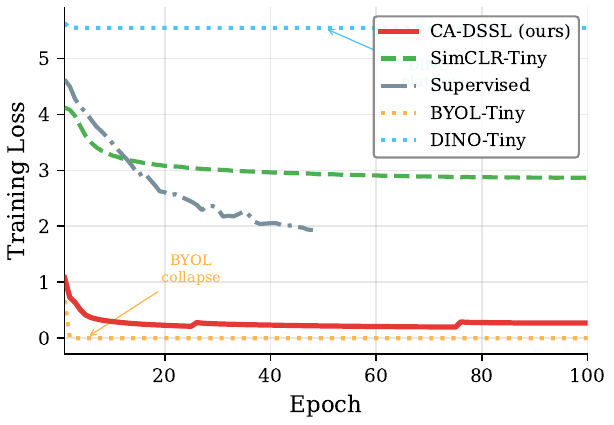}
\caption{\textbf{Pretraining loss curves} (CIFAR-100, 100 epochs). BYOL-Tiny collapses to zero loss immediately; DINO-Tiny plateaus without learning. \method{} converges to the lowest meaningful loss. Note: losses are not directly comparable across methods (different objectives).}
\label{fig:training_curves}
\end{figure}

\section{Failure Analysis}
\label{app:failure}

\paragraph{BYOL and DINO-Tiny collapse.}
Both methods collapse to near-random linear-probe accuracy
($\sim\!5\%$ on CIFAR-100). BYOL-Tiny's loss drops to zero within the
first few epochs, indicating representation collapse: the EMA teacher and
online student converge to a trivial constant output. DINO-Tiny's loss
plateaus immediately, suggesting that centering and sharpening cannot
produce meaningful self-distillation targets from a 396K-parameter backbone.
We trained both methods with the default hyperparameters used for their
respective large-model versions (BYOL: $\text{lr}{=}10^{-3}$, EMA decay
$0.996$; DINO: $\text{lr}{=}10^{-3}$, centering momentum $0.9$).
A limited sweep over learning rates ($\{5 \times 10^{-4}, 10^{-3},
3 \times 10^{-3}\}$) and EMA decay ($\{0.99, 0.996, 0.999\}$) did not
resolve the collapse for either method. We attribute BYOL's failure to EMA
instability with very few parameters: the target network cannot diverge
meaningfully from the online network. DINO's failure stems from the
representation bottleneck: $\sim$112 final-stage channels cannot support the
diversity of cluster assignments that centering requires.

\paragraph{CLIP teacher underperforming DINO teacher.}
The CLIP ViT-B/32 teacher (88M params) produces a $\sim\!3$~pp drop in linear
probe accuracy relative to DINO ViT-S/16 (22M), despite being the larger
model. CLIP's image encoder exposes only a global image-text-aligned CLS
token; it does not provide spatially structured intermediate features.
This effectively disables multi-scale spatial distillation
($\mathcal{L}_\text{ms}$), reducing \method{} to CLS-only distillation
against a teacher whose representation space is optimized for image-text
alignment rather than image-image similarity. DINO's patch-level features
and self-supervised spatial structure are better suited to producing
informative multi-scale targets for a tiny student.

\section{Scaling Analysis: ImageNet-100 Pretraining}
\label{app:imagenet100}

To test whether \method{}'s advantages persist with larger pretraining data, we train all methods on ImageNet-100 (100 classes, 130K images from ImageNet-1K) and evaluate via CIFAR-100 linear probe (cross-dataset transfer). \Cref{tab:imagenet100} reports results.

\begin{table}[h]
\centering
\caption{\textbf{ImageNet-100 pretraining $\rightarrow$ CIFAR-100 linear probe} (cross-dataset transfer). Single seed (42) except \method{} (3-seed mean $\pm$ std).}
\label{tab:imagenet100}
\small
\begin{tabular}{lcc}
\toprule
Method & IN100 $\rightarrow$ CIFAR-100 & CIFAR-100 $\rightarrow$ CIFAR-100 \\
\midrule
\method{} ($\mathcal{L}_\text{cls} {+} \mathcal{L}_\text{ms}$) & $24.0 \pm 0.7$ & $\mathbf{62.7 \pm 0.5}$ \\
\method{} ($+ \mathcal{L}_\text{reg}$) & $23.7$ & $60.5 \pm 0.4$ \\
SEED & $\mathbf{44.1}$ & $61.7 \pm 0.4$ \\
SimCLR-Tiny & $35.2$ & $44.5 \pm 0.9$ \\
Supervised & $40.8$ & $66.7 \pm 0.3$ \\
\bottomrule
\end{tabular}
\end{table}

\paragraph{Finding: \method{} does not scale to larger pretraining data.}
On ImageNet-100, SEED outperforms \method{} by $+20$~pp ($44.1\%$ vs.\ $24.0\%$), reversing the ordering seen on CIFAR-100 where \method{} leads. Enabling $\mathcal{L}_\text{reg}$ does not help ($23.7\%$), contra our hypothesis in \Cref{sec:experiments}. We attribute this to the capacity-proportional projection head: with only 328K parameters ($0.8\times$ backbone), it cannot encode the richer feature diversity present in 130K ImageNet images. SEED's 3.15M MLP head ($8\times$ backbone) has sufficient capacity to exploit the larger dataset.

\paragraph{Implication.}
\method{} is specifically designed for the \emph{small-data MCU regime} where (i) pretraining corpora are limited (tens of thousands of images), (ii) training-time memory is constrained, and (iii) deployment efficiency dominates. When abundant pretraining data is available and training is performed on GPUs with no memory constraints, a SEED-style baseline with a larger projection head is preferable. This is a deliberate design trade-off, not a failure: the capacity-proportional head sacrifices scalability for deployment-relevant parameter efficiency.

\section{MCU Deployment Measurements}
\label{app:mcu}

\Cref{tab:mcu} reports on-device benchmarks for the deployed
MobileNetV2-0.35$\times$ backbone across three MCU platforms. Because
\method{} modifies only pretraining (not architecture or weights count),
these numbers are identical regardless of which pretraining method was
used; they document the inference-time cost of the deployed artefact.

\begin{table}[h]
\centering
\caption{\textbf{MCU deployment measurements} for the INT8-quantized
$\sim\!378$~KB MobileNetV2-0.35$\times$ backbone. Pretraining method does
not affect any of these values---the deployed binary is the same.}
\label{tab:mcu}
\small
\begin{tabular}{lccc}
\toprule
 & STM32H7 & ESP32-S3 & MAX78000 \\
\midrule
Flash (KB)   & 268  & 484  & 51  \\
SRAM (KB)    & 299  & 294  & --- \\
Latency      & 6.75s & 26.1s & 9.6ms$^\dagger$ \\
\midrule
Model (INT8) & \multicolumn{3}{c}{377.6 KB (386K params)} \\
\bottomrule
\end{tabular}
\vspace{0.5em}
\raggedright\footnotesize{$^\dagger$The MAX78000 CNN accelerator processes
only the backbone convolutions; full-pipeline latency would be higher.}
\end{table}

\section{Dataset Details and Licenses}
\label{app:datasets}

\Cref{tab:datasets} summarises the datasets used in this paper.
All datasets are publicly available and used under their respective licenses.

\begin{table}[h]
\centering
\caption{\textbf{Datasets used.}}
\label{tab:datasets}
\small
\begin{tabular}{llll}
\toprule
Dataset & Role & License & Source \\
\midrule
CIFAR-100 & Pretraining + eval & MIT & Krizhevsky (2009) \\
ImageNet-100 & Pretraining (scaling) & Non-commercial & Deng et al.\ (2009) \\
Flowers102 & Transfer probe & Non-commercial & Nilsback \& Zisserman (2008) \\
Food101 & Transfer probe & Research-only & Bossard et al.\ (2014) \\
Core50 & Transfer probe & CC BY 4.0 & Lomonaco \& Maltoni (2017) \\
VOC 2007/2012 & Detection transfer & Non-commercial & Everingham et al.\ (2010) \\
COCO 70+10 & Detection transfer & CC BY 4.0 & Lin et al.\ (2014) \\
\bottomrule
\end{tabular}
\end{table}

\end{document}